\documentclass[onefignum,onetabnum]{siamonline190516}

\usepackage{lipsum}
\usepackage{amsfonts}
\usepackage{graphicx}
\usepackage{epstopdf}
\usepackage{algorithmic}
\usepackage{subcaption}
\ifpdf
  \DeclareGraphicsExtensions{.eps,.pdf,.png,.jpg}
\else
  \DeclareGraphicsExtensions{.eps}
\fi

\usepackage{booktabs}

\usepackage{enumitem}
\setlist[enumerate]{leftmargin=.5in}
\setlist[itemize]{leftmargin=.5in}

\newsiamremark{remark}{Remark}
\newsiamremark{hypothesis}{Hypothesis}
\crefname{hypothesis}{Hypothesis}{Hypotheses}
\newsiamthm{claim}{Claim}

\headers{Topological Data Analysis for Anomaly Detection in Host-Based Logs}{Thomas Davies}

\title{Topological Data Analysis for Anomaly Detection in Host-Based Logs}

\author{Thomas Davies\thanks{The Alan Turing Institute, The British Library, London, UK.
  (\email{tdavies@turing.ac.uk}).}
}
\usepackage{amsopn}

\ifpdf
\hypersetup{
  pdftitle={TDA for Anomaly Detection in Host-Based Logs},
  pdfauthor={Thomas Davies}
}
\fi

\begin{document}

\maketitle

\begin{abstract}
Topological Data Analysis (TDA) gives practioners the ability to analyse the global structure of cybersecurity data. We use TDA for anomaly detection in host-based logs collected with the open-source Logging Made Easy (LME) project. We present an approach that builds a filtration of simplicial complexes directly from Windows logs, enabling analysis of their intrinsic structure using topological tools. We compare the efficacy of persistent homology and the spectrum of graph and hypergraph Laplacians as feature vectors against a standard log embedding that counts events, and find that topological and spectral embeddings of computer logs contain discriminative information for classifying anomalous logs that is complementary to standard embeddings. We end by discussing the potential for our methods to be used as part of an explainable framework for anomaly detection.

\end{abstract}



\section{Introduction}

Topological Data Analysis (TDA) is a collection of techniques that concisely summarise the topology of a dataset using tools from algebraic topology. Considering the topology of a dataset, which is roughly equivalent to its global shape, gives analysts new ways to approach analysing data. This is particularly applicable to cybersecurity, where individual indicators of anomalous behaviour on a system may not be conclusive, but taken together provide strong evidence of compromise. We analyse host-based logs using TDA and find that it provides comparable results to more traditional count-based embeddings, demonstrating that feature vectors computed using the topological structure provide discriminative information for detecting anomalous logs with downstream classifiers.

Previous work analysing cybersecurity data using TDA differs from ours in two ways. Firstly, it focuses entirely on network logs, whereas we analyse host-based logs. Secondly, we directly embed logs into the structures required for TDA, capturing their intrinsic structure whilst avoiding the computationally expensive construction that is often used instead. This technique can also be used to embed logs into graphs and hypergraphs, so we analyse the spectrum of the graph and hypergraph Laplacians as alternate feature vectors. We find that they are also effective for anomaly detection, implying that even though we lose explicit information about the logs when we embed them, the structure that we capture using topological and spectral techniques contains discriminative information on whether the logs are anomalous. Furthermore, we find that the combination of a count feature vector and a spectral feature vector equals or betters any other method, suggesting that structural feature vectors can provide complementary information to more traditional techniques.

\section{Background}

We start by briefly introducing persistence diagrams, the main output of TDA, and simplicial complexes: the data structure you need to compute them. We use that knowledge to review previous work on persistence diagrams applied to cybersecurity data.

\subsection{Persistence diagrams}

Persistence diagrams give a concise summary of the global structure of a dataset. They cannot be computed directly from data living in $\mathbb{R}^n$, but are instead computed from a \textit{filtration of simplicial complexes}. An (abstract) simplicial complex is similar to a graph: it consists of a collection of 0-simplices (nodes) and 1-simplices (edges), but also higher-order equivalents: 2-simplices (triangles, or a clique of three nodes), and $k$-simplices -- cliques of $k+1$ nodes. A simplicial complex $K$ is a collection of simplices such that if a $k$-simplex is in $K$ then so are all of its constituent simplices. For example, if a 2-simplex (triangle) is in $K$ then so are its constituent $0$ and $1$-simplices (nodes and edges). The intersection of any two simplices in $K$ must also be a simplex in $K$. A filtration of simplicial complexes is a collection of simplicial complexes $(K_i)_{i \in \mathbb{R}}$ such that $K_i \subseteq K_j$ for any $i \leq j$. In computational topology such a filtration is often built using the Vietoris-Rips complex \cite{rips}, a construction which adds $k+1$ points to the complex as a $k$-simplex when they are pairwise at least $\epsilon$ distance away from each other. See Appendix A for further details on this computation.

Given a filtration of simplicial complexes $(K_i)$ we can compute the persistence diagram: a multiset of points in the extended plane that concisely represents the topology of the dataset. Each point in the $0$-persistence diagram represents a connected component, in the $1$-persistence diagram represents a hole, in the $2$-persistence diagram represents a void (e.g., the inside of a football), and so on. As simplices are added to the filtration, topological features are created: a hole may be born within the filtration when simplices are added at time $i$. More simplices that fill in the hole could be added at time $j$. We say that this hole is a topological feature that is born at time $i$, dies at time $j$, and persists for $j-i$. It would be represented by a point in the $1$-persistence diagram at $(i,j)$. In order to make persistence diagrams more amenable to standard machine learning workflows, we embed them into a vector. The most common way to do this (and the technique we use) is to embed the persistence diagram into a persistence image \cite{persimages}. This transforms the birth-death axes of the diagram to birth-persistence, puts a Gaussian over each point, and integrates over a grid to create a vector of fixed size which can be used as an input to most machine learning algorithms. Further details are in Appendix A.

\begin{figure}[t!]
\centering

\includegraphics[width=0.99\textwidth]{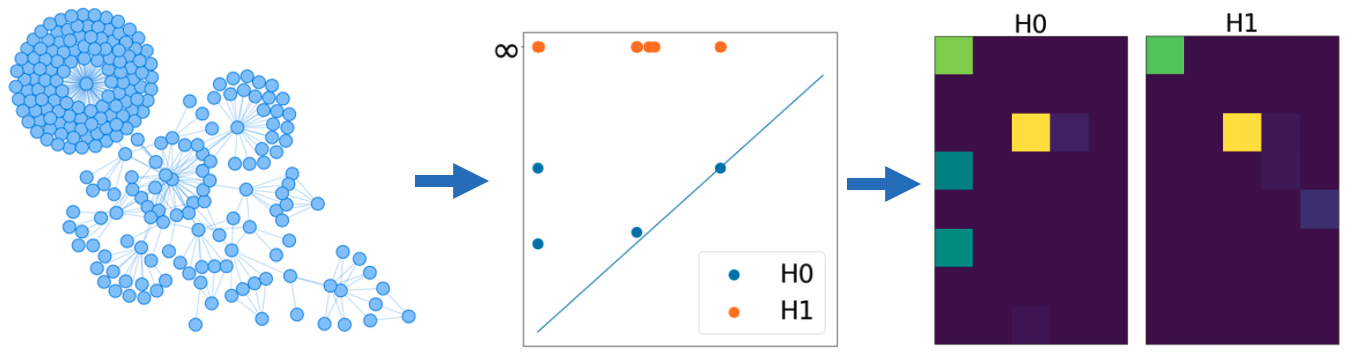}

\caption{An example of Windows logs that have been embedded into a simplicial complex, along with the corresponding persistence diagram and images.}
\vspace{-1mm}
\end{figure}

\subsection{Anomaly detection and host-based logs}

In cybersecurity it is desirable to seek out anomalous behaviour. That is, any behaviour which is out of the ordinary and can indicate that there is something happening on a computer that shouldn't be; for example, malware executing. This paper focuses in particular on detecting anomalies in host-based logs - collections of events on a computer, or host, that have been collected by a service running on that computer. By considering a collection of host-based logs together, you may be able to discern whether or not something anomalous is taking place on that host. The host-based logs that we analyse are Windows System Monitor (Sysmon) logs captured by the open-source \href{https://github.com/ukncsc/lme}{Logging Made Easy} project. Sysmon logs can capture a variety of events taking place on a computer. Depending on the specific event they will include information like the unique process identifier, child processes, command line prompts, network information such as IP addresses and ports, and more. A standard way of embedding logs into real-valued vectors for analysis is the count vectorisation strategy, which was first suggested by Winding, Wright, and Chapple \cite{vectorisation}. This vector consists of a count of the number of each type of log, along with a count of the number of unique values appearing for select attributes of the logs.

\subsection{Previous work}

Persistent homology has been applied to cybersecurity data several times before. Bruillard, Nowak, and Purvine \cite{anom_ph} vectorised NetFlow data by creating a count vector of the number of network packets flowing between IPs of interest over a certain time frame. They compute the corresponding persistence diagram for the vectors. If the distance between the persistence diagram and a persistence diagram they know corresponds to non-anomalous activity is large, then the new data is likely to be anomalous. They showed they can detect DDOS attacks and port scans using this technique. Postol et al. \cite{postol2019time} similarly embedded network logs from internet of things (IoT) devices into count vectors then computed the Vietoris-Rips complex, the persistence diagram, and an embedding of the diagram to classify types of IoT devices from the network traffic. Collins et al. \cite{passive} extended on this work by building a simplicial complex from the packet data. One central node represents the IoT device of interest, and other nodes represent packets that are going to it. Packet nodes are connected sequentially and to the central IoT device node, then a filtration is induced by inter-packet arrival time. By computing the persistence image for each filtration and classifying using a CNN they are able to predict the IoT device even when the traffic is encrypted. Our embeddings are very similar to that of Aksoy, Purvine, and Young \cite{aksoy2020directional}, who embed network data into graphs, rather than host-based data into simplicial complexes. If the reader is interested, we have recently written a review paper on TDA for cybersecurity \cite{davies}.

\begin{figure*}[t!]
\centering
\includegraphics[width=0.99\textwidth]{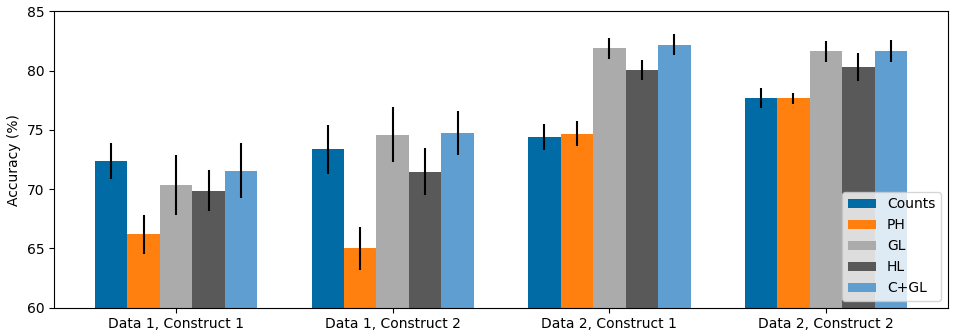}
\caption{Anomalous logs classification accuracy across all datasets, contructions, and feature embeddings. Results are presented for the counts baseline vector (C), persistent homology with the 0th and 1st dimensions (PH), the graph Laplacian (GL), and the hypergraph Laplacian (HL), as well as using both the counts vector and graph Laplacian (C+GL). The classifier is a 10-fold cross-validated random forest. Precision, recall, and F1 are available for all experiments in Tables 1 and 2.}
\vspace{-1mm}
\end{figure*}

\section{Methodology}

We build filtrations of simplicial complexes from Windows Sysmon logs. Given a collection of logs that we wish to classify as either benign or anomalous we build our filtration of simplicial complexes as follows. The 0-simplices are unique identifiers within the logs such as process IDs, file names, IP addresses and ports. They enter the filtration at the time they first appear in the logs. We add 1-simplices when two 0-simplices interact with each other in a log entry. For example, when a process launches another process, or there is a network connection between two IP addresses. The 1-simplex enters the filtration at the time of the corresponding log entry. We experimented with adding induced 2-simplices, however we found that it makes a negligible difference to performance (see Table 3 in Appendix B). We compute the persistence diagram of the logs, then the persistence images.\footnote{Computed using \href{https://mrzv.org/software/dionysus2/}{Dionysus} and \href{https://persim.scikit-tda.org/en/latest/}{Persim}.} An example of this is shown in Figure 1. We experimented with different resolutions for the persistence images, and found that it did not make much difference (Table 4, Appendix B).

We also considered the spectrum of the graph Laplacian and the hypergraph Laplacian as features.\footnote{Computed using \href{https://scipy.org/}{SciPy} \cite{2020SciPy-NMeth} and \href{https://pnnl.github.io/HyperNetX/build/index.html}{HyperNetX} \cite{joslyn2020hypernetwork}.} By considering the 0 and 1-simplices of the final simplicial complex in the filtration we obtain a graph. We compute the spectrum of its Laplacian, which we use as a feature vector. We can similarly obtain a hypergraph Laplacian by building a hypergraph from log entries, and using its spectrum as a feature vector. Note the difference between a simplicial complex and a hypergraph: a hypergraph doesn't require each sub-edge to be in the hypergraph, so is a more general version of a simplicial complex. By adding hyperedges consisting of log entires we can build a hypergraph. To give an example, one hyperedge could be a process ID that launched a network event connecting one IP and port to another IP and port, leading to a hyperedge consisting of five nodes. Finally, we implemented the count vectorisation strategy discussed in Section 2.2 as our baseline \cite{vectorisation}. This vector is commonly used in the literature for automated anomaly detection in logs, so is a suitable comparison for our techniques \cite{enterprise, anomex1}. We refer to this baseline as `counts'.

We use a 10-fold cross validated random forest as the classifier for each feature vector. Additional details on all methods, including parameters for the random forest and a discussion of test set stratification, are available in Appendix B.

{ 
\begin{table}[h!]
\centering
\caption{Experimental results with dataset 1. Classification using a 10-fold cross validated random forest. Results are presented for the counts baseline vector (Counts), persistent homology with the 0th and 1st dimensions (PH), the graph Laplacian, and the hypergraph Laplacian, as well as using both the counts vector and graph Laplacian.}
\small
\begin{tabular}{llllll} \toprule
& \multicolumn{5}{c}{\textbf{Construction 1}} \\\cmidrule(lr){2-6}
   & Counts & PH (H0/H1) & Graph Lap. & Hyper. Lap. & Counts + GL \\ \midrule
Accuracy  &    $72.37 \pm 1.49$      & $66.20 \pm 1.65$ &  $70.37  \pm 2.53$    & $69.89 \pm 1.73$              & $71.55 \pm 2.31$                                     \\
Precision & $74.72 \pm 1.15$ & $68.85 \pm 1.77$         &  $72.07 \pm 2.22$                                           &   $72.32 \pm 1.58$                 &  $73.70 \pm 2.10$                  \\
Recall    &   $70.92 \pm 2.90$       &                $64.58 \pm 1.72$             &     $69.95 \pm 3.51$          &        $68.35 \pm 3.10 $            & $70.44 \pm 3.25$                   \\
F1        &    $72.53 \pm 1.84$      &      $66.56 \pm 1.59$                       &     $70.83 \pm 2.73$          &    $70.00 \pm 2.05$                &  $71.83 \pm 2.50$
\\\midrule
& \multicolumn{5}{c}{\textbf{Construction 2}} \\\cmidrule(lr){2-6}
   & Counts & PH (H0/H1) & Graph Lap. & Hyper. Lap. & Counts + GL \\ \midrule
Accuracy  &   $73.36 \pm 2.07$       & $65.01 \pm 1.83$ &  $74.61 \pm 2.33$    & $71.46 \pm 1.98$              & $74.74 \pm 1.88$                                       \\
Precision &   $76.53 \pm 1.93$       &        $67.97 \pm 1.90$                     &      $78.45 \pm 2.07$         &      $73.95 \pm 1.97$              &      $79.03 \pm 1.70$              \\
Recall    &     $70.32 \pm 2.81$     &          $62.31 \pm 1.94$                   &     $70.45 \pm 3.48$          &          $69.26 \pm 2.76$          &       $69.97 \pm 2.90$             \\
F1        &    $73.17 \pm 2.26$      &      $64.96 \pm 1.84$                       &    $73.97 \pm 2.71$           &      $71.38 \pm 2.22$              &  $74.04 \pm 2.21$

\\\bottomrule
\end{tabular}
\end{table}
}

{ 
\begin{table}[h!]
\centering
\caption{Experimental results with dataset 2. The classifier and methods are the same as Table 1.}
\small
\begin{tabular}{llllll} \toprule
& \multicolumn{5}{c}{\textbf{Construction 1}} \\\cmidrule(lr){2-6}
   & Counts & PH (H0/H1) & Graph Lap. & Hyper. Lap. & Counts + GL \\ \midrule
Accuracy  & $74.44 \pm 1.10$ & $74.69 \pm 1.05$ & $81.88 \pm 0.89$ & $80.05 \pm 0.83$ & $82.19 \pm 0.88$ \\
Precision & $75.91 \pm 1.01$ & $74.22 \pm 0.85$ & $83.12 \pm 0.77$ & $81.51 \pm 0.80$ & $83.04 \pm 0.97$ \\
Recall    & $71.62 \pm 1.83$ & $75.49 \pm 1.73$ & $80.00 \pm 1.43$ & $77.71 \pm 1.37$ & $81.00 \pm 1.34$ \\
F1        & $73.61 \pm 1.27$ & $74.80 \pm 1.22$ & $81.48 \pm 0.98$ & $79.51 \pm 0.95$ & $81.94 \pm 0.93$ \\ \midrule
& \multicolumn{5}{c}{\textbf{Construction 2}} \\\cmidrule(lr){2-6}
   & Counts & PH (H0/H1) & Graph Lap. & Hyper. Lap. & Counts + GL \\ \midrule
Accuracy  & $77.69 \pm 0.87$ & $77.69 \pm 0.47$ & $81.62 \pm 0.88$ & $80.31 \pm 1.22$ & $81.69 \pm 0.93$ \\
Precision & $79.52 \pm 0.76$ & $76.80 \pm 0.81$ & $81.63 \pm 0.79$ & $81.38 \pm 1.22$ & $81.73 \pm 1.06$  \\
Recall    & $74.62 \pm 1.61$ & $79.49 \pm 0.84$ & $81.72 \pm 1.37$ & $78.62 \pm1.48$ & $81.75 \pm 1.24$ \\
F1        & $76.91 \pm 1.02$ & $78.07 \pm 0.43$ & $81.59 \pm 0.96$ & $79.95 \pm 1.28$ & $81.69 \pm 0.94$ \\ \bottomrule
\end{tabular}
\end{table}
}

\subsection{Datasets} 

We run experiments on two different private datasets. Each dataset consists of approximately 1600 runs (see Table 5, Appendix B) of a Windows machine, where each run is around three minutes of activity that has been logged using the open-source \href{https://github.com/ukncsc/lme}{Logging Made Easy} project. Each run consists of either baseline or anomalous activity on the system. In Dataset 1 the activity is the execution of cleanware (baseline) or malware (anomalous). Each piece of software executed in Dataset 1 is unique to the run. In comparison, Dataset 2 attempts to emulate an attack on the system. There are 160 baseline runs and 160 `attack’ runs, consisting of a mixture of fileless attacks (using legitimate tools already on the system to enact malicious behaviour), attacks using files pre-installed on the machine (in the assumption the malware has already reached the machine through a phishing attack or similar), or attacks using malware that is downloaded from a remote source before execution. Each run is present in the dataset unobfuscated and with four obfuscation techniques (such as Base64 and Caeser cipher encodings) applied to key descriptors in the logs. Therefore Dataset 2 has a larger variety of (sometimes obfuscated) attack types present in the anomalous data, but there are fewer unique attacks.

\section{Main results}

We compare four different techniques: our counts baseline, persistent homology (consisting of the 0 and 1-persistence images), the spectrum of the graph Laplacian, and the spectrum of the hypergraph Laplacian. In addition we ran all experiments using two different sets of logs, which we refer to as constructions. In Construction 1 we use process creation and network event logs, and in Construction 2 we use process creation, process termination, file creation, and network event logs. For each experiment we get the average accuracy, precision, recall, and F1 (with error bounds) from a 10-fold cross validated random forest. Figure 2 shows the accuracy, and all metrics are available in Tables 1 and 2. It is not a surprise that the more logs we add, the better our models tend to perform: in Dataset 1, Construction 2 consistently outperforms Construction 1 by 4-5\%. The difference is smaller in Dataset 2, but is not insignificant in some metrics and methods. The compromise is that adding more logs increases the computational complexity: if you wish to run these methods as close to real-time as possible, you may choose to use fewer types of logs, sacrificing some performance in return for faster computation times.

In Dataset 1, we find that counts outperforms persistent homology by approximately 6\%, and is marginally better than the graph and hypergraph Laplacians. In Dataset 2, counts equals persistent homology, and is outperformed by the graph and hypergraph Laplacians. The fact that the topological and spectral techniques that we investigate are able to perform at a level that rivals or even outperforms a standard baseline should be considered the primary result of this paper. Unlike count vectors, which use explicit details about what different event logs are, the topological and spectral feature vectors only use the global structure that is intrinsic to the logs. The fact that our techniques are able to rival the counts baseline means that \textit{just the structural information is somehow fingerprinting the anomalous behaviour.} There was no guarantee that this would be the case. The higher overall results in Dataset 2 are likely because our methods won't be affected by the obfuscation methods applied, so its likely there is some overlap between train and test sets (modulo the obfuscation methods). Note that this applies to every technique, so the comparison between techniques remains valuable.

We further hypothesised that the counts baseline and topological/spectral techniques were learning different information. To test this we created a new feature vector that appended the counts vector to the best performing of the structural vectors: the graph Laplacian. The feature vector created by combining the two either rivalled or outperformed almost every method in almost every metric, supporting our hypothesis.

\begin{figure*}[]
\centering
\includegraphics[width=0.99\textwidth]{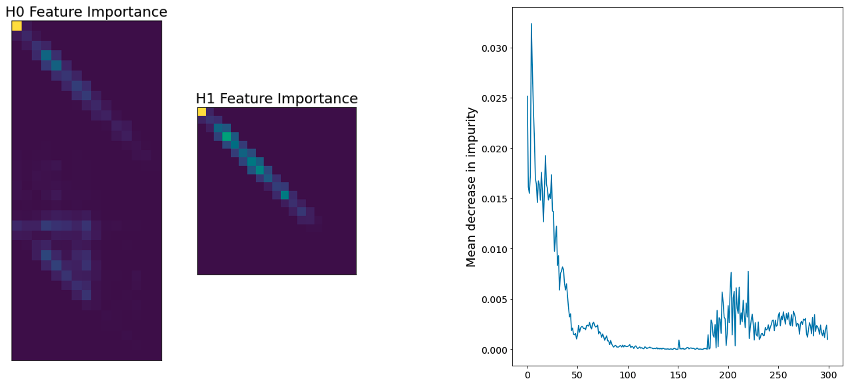}
\caption{The MDI feature importance of a persistence image (left), and the feature importance of the spectrum of the graph Laplacian with the $x$-axis sorted by decreasing size of the eigenvalue (right).}
\vspace{-1mm}
\end{figure*}

\section{Explainability} We ran feature importance analysis on our models using the Mean Decrease in Impurity (MDI) feature importance technique for random forests \cite{scornet2020trees}. For the persistent homology features we get a ranking over the persistence images, shown on the left of Figure 3, which tell us that topological features that are born at the start of the filtration and persist throughout it are most important. In dataset 1 and part of dataset 2 each experimental run starts with the execution of the cleanware/malware, suggesting that our methods are generally picking up on the initial execution of the software. What exactly these topological features correspond to in the logs requires further investigation to fully understand. For the graph Laplacian we investigated the feature importance relative to the size of the eigenvalue (see the right of Figure 3). We see that generally the largest eigenvalues are most discriminative, although there is a spike towards the smaller eigenvalues that we hypothesise is due to the importance of small eigenvalues in characterising the structure of graphs \cite{nica2016brief}. As each element in an eigenvector corresponds to a node in the graph we can plot those values on the nodes, leading to a visualisation that could be used to flag nodes (and therefore logs) of interest.

\section{Conclusions and future work}
\label{sec:conclusions}

We have demonstrated that the intrinsic structure of host-based logs, as captured by persistence images and the spectrum of graph and hypergraph Laplacians, contains discriminative information about whether or not the logs are anomalous. Although this information has been demonstrated using synthetic data with balanced classes, the evidence that structural feature vectors contain useful information can hopefully be ported to real-life scenarios where malign activity is far outweighed by benign activity. Expanding on our initial evidence that structural information is complementary to more traditional techniques and better understanding the explainability aspect of this research is future work that would lead to better understanding of these methods; investigating how our feature vectors perform on separate attack types in Dataset 2 would also give more insight. In summary, we have provided evidence that topological feature vectors can be used for anomaly detection; future research should investigate what exactly within the logs these techniques are identifying.

\newpage
\appendix

\section{Additional background on TDA}

Given points in $\mathbb{R}^n$, we create the Vietoris-Rips complex $K_\epsilon$ by adding $k$ points to the complex as a $(k+1)$-simplex if they are pairwise within $\epsilon$ of each other (Figure 3). Note that as increasing $\epsilon$ will only add new simplices to the complex, the collection of Vietoris Rips complexes $(K_\epsilon)_{\epsilon \in \mathbb{R}_0^+}$ is a filtration of simplicial complexes, as required. An example of this is shown in Figure 3. Although this is the most common way to build a filtration from data in TDA, it is not the only way. In fact, in this paper we build our filtration directly from computer logs, bypassing the expensive computation of the Vietoris-Rips complex and building simplicial complexes using the structure inherent to the data we're working with. Figure 4 shows an example of how a persistence diagram is turned into a persistence image.

\begin{figure*}[h!]
\centering
\includegraphics[width=0.95\textwidth]{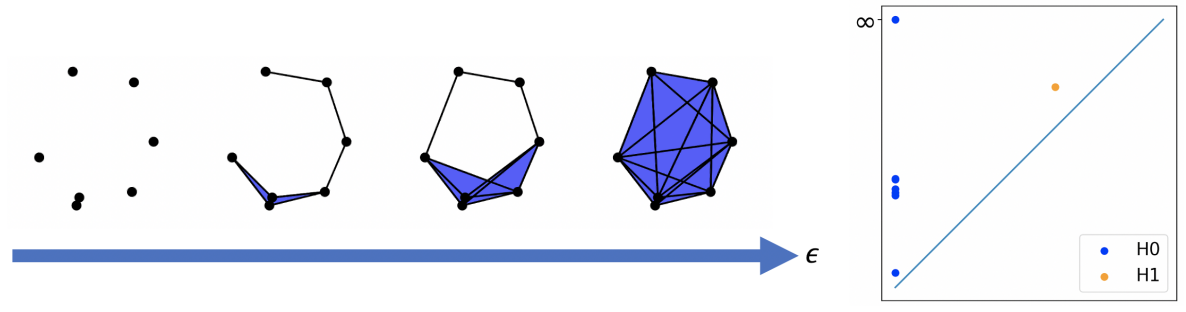}
\caption{An example of a filtration of simplicial complexes with its corresponding persistence diagram. In particular, this is a Vietoris-Rips filtration. This figure is taken from our recent review on TDA for cybersecurity \cite{davies}.}
\end{figure*}

\begin{figure*}[h!]
\centering
\includegraphics[width=0.8\textwidth]{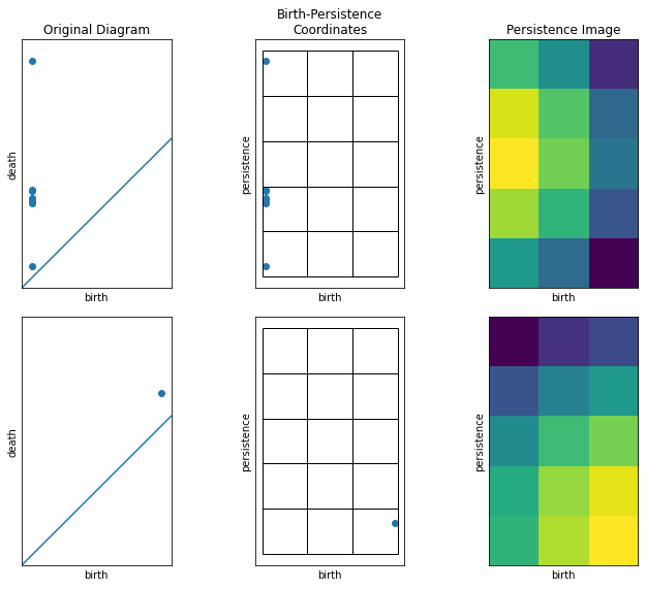}
\caption{An example of how the persistence diagram in Figure 3 is split into the 0 and 1 persistence diagram and the persistence images are computed.}
\end{figure*}

\section{Experimental details}

Additional statistics on the datasets are available in Table 5. We also provide implementation details for each method below.

\subsection{Counts baseline} We create a vector for each experimental run by counting the number of each event (corresponding to the logs included in the particular construction), followed by counting the number of unique identifiers corresponding to logs in the construction. For example, if network events are included, then we will count the number of logs of network events, and the number of unique destination IPs and destination ports.

{
\begin{table}[]
\centering
\caption{Investigation of adding induced 2-simplices, using dataset 1, construction 1, and 0/1-persistence diagrams. We found that there was no real change when adding induced 2-simplices. We hypothesise this is because most holes were never destroyed, even when adding the induced 2-simplices. As this was computationally costly we did not add induced 2-simplices in the rest of the experiments.}
\small
\begin{tabular}{lll} \toprule
          & No 2-simplices & Induced 2-simplices \\ \midrule
Accuracy  & $66.20 \pm 1.65$ & $66.14 \pm 1.49$ \\
Precision & $68.85 \pm 1.77$ & $68.56 \pm 1.54$ \\
Recall    & $64.58 \pm 1.72$ & $64.34 \pm 1.98$ \\
F1        & $66.56 \pm 1.59$ & $66.34 \pm 1.61$ \\ \bottomrule
\end{tabular}
\end{table}
}

{
\begin{table}[]
\centering
\caption{Investigation of persistence image pixel size effect on performance of classifier, using dataset 1, construction 1, and 0/1-persistence diagrams. We found that there was no real change when using different sizes of pixels. Based on this we selected a pixel size of 20 as it offered a compromise between resolution and the dimensionality of the persistence image vector.}
\small
\begin{tabular}{lllllll} \toprule
 Pixel size  & 5 & 10 & 15 & 20 & 50 & 100 \\ \midrule
Accuracy  & $63.95 \pm 2.06$ & $65.07 \pm 2.19$ & $65.58 \pm 1.89$ & $64.26 \pm 2.22$ & $64.51 \pm 1.87$ & $61.96 \pm 1.77$ \\
Precision & $66.15 \pm 2.05$ & $67.10 \pm 2.19$ & $67.53 \pm 1.98$ & $66.31 \pm 2.12$ & $66.54 \pm 1.78$ & $63.83 \pm 1.62$ \\
Recall    & $63.37 \pm 2.54$ & $64.81 \pm 2.59$ & $65.77 \pm 2.10$ & $63.37 \pm 3.04$ & $64.10 \pm 2.23$ & $61.93 \pm 2.75$ \\
F1        & $64.59 \pm 2.11$ & $65.82 \pm 2.24$ & $66.54 \pm 1.85$ & $64.68 \pm 2.48$ & $65.23 \pm 1.94$ & $62.71 \pm 2.05$ \\ \bottomrule
\end{tabular}
\end{table}
}

{
\begin{table}[h!]
\centering
\caption{The number of examples of malware and cleanware in each dataset.}
\small
\begin{tabular}{lll} \toprule
          & Dataset 1 & Dataset 2 \\ \midrule
Cleanware  & 768  & 800\\
Malware    & 841  & 800 \\ \midrule
Total      & 1609 & 1600 \\ \bottomrule
\end{tabular}
\end{table}
}

\subsection{Persistent homology} The method which we use to build a filtration of simplicial complexes from logs is described in Section 3 of the main paper. We compute the persistence diagram using Dionysus. Any points at infinity we change to a value that is much larger than the maximum value in the filtration so that we are able to compute the persistence image. For example, if the largest death value in the filtration was 200, we would change all points at infinity in the persistence diagram to have death time 500. We compute the persistence image using the default kernel and weighting from Persim. We investigated the effect of changing the pixel size, and found that it was negligible: details are in Table 4. We also investigated the effect of adding induced 2-simplices to our simplicial complexes and found that it made very little difference (Table 3). We found that generally 1-features weren't being killed when adding 2-simplices. This, coupled with the fact that adding induced 2-simplices is very computationally intensive, meant that we generally didn't add induced 2-simplices when running our experiments.

\subsection{Graph Laplacian} We computed the graph Laplacian of graphs that were simply the 0 and 1-simplices of the simplicial complexes we'd built. As the size of the graph Laplacian, and therefore the number of eigenvalues, depends on the size of the graph, we fixed a set size for the graph spectrum feature vector for each experiment. We averaged the number of eigenvalues across all runs for each experiment, and set the feature vector length as slightly above the average. We ordered the eigenvalues by size (largest first), truncating if there were too many eigenvalues and zero-padding if there were too few.

\subsection{Hypergraph Laplacian} We built our hypergraphs using the HyperNetX package. We used the logs according to the experiment we were running, but in order to exploit hyperedges we added them to the hypergraph slightly differently. If there were more than two nodes in a log entry, we would add all of them to one hyperedge, instead of adding them as 1-simplices pairwise. We set the length of the feature vector based on the average number of eigenvalues across each experiment in exactly the same way as for the graph Laplacian.

\subsection{Classifier}

We use scikit-learn's RandomForestClassifier with default settings - the specific parameters are shown in Table 7. We train and score using scikit-learn's cross\_validate over 10 random folds. Note that we don't stratify the test sets for Dataset 2 based on the obfuscation methods. This likely contributes to the increase in overall performance for Dataset 2, although this is true for every method, so the comparisons across methods remain reliable.

{
\begin{table}[]
\centering
\caption{Parameters for the scikit-learn random forest classifier used for all experiments.}
\small
\begin{tabular}{ll} \toprule
Parameter          & Value \\ \midrule
bootstrap  & True \\
ccp\_alph & 0.0 \\
class\_weight    & None  \\
criterion & `gini' \\
max\_depth & None  \\
max\_features & `auto' \\
max\_leaf\_nodes & None \\
max\_samples & None \\
min\_impurity\_decrease & 0.0 \\
min\_samples\_leaf & 1 \\
min\_samples\_split & 2 \\
min\_weight\_fraction\_leaf & 0.0 \\
n\_estimators  & 100 \\
n\_jobs & None \\
oob\_score & False \\
random\_state & 0 \\
verbose & 0 \\
warm\_start & False \\ \bottomrule
\end{tabular}
\end{table}
}

\section*{Acknowledgments}
This research was supported by the Defence and Security programme at the Alan Turing Institute, funded by the UK Government.

\bibliographystyle{siamplain}
\bibliography{tda_for_cyber}

\end{document}